\documentclass[letterpaper, 10 pt, conference]{ieeeconf}  

\IEEEoverridecommandlockouts                              

\usepackage{lipsum}
\usepackage{graphicx}
\usepackage{graphics}
\usepackage{amsfonts}
\usepackage{siunitx}
\usepackage{float}
\usepackage[hidelinks]{hyperref}
\usepackage{epstopdf}
\usepackage{enumerate}
\usepackage{cite}
\usepackage{xcolor}
\usepackage{amsmath}
\usepackage{bm}
\graphicspath{{figures/}}
\usepackage{balance}

\newcommand{\R}{\mathbb{R}}
\newcommand{\norm}[1]{\left\lVert#1\right\rVert}

\newcommand{\SE}{SE(3)}
\def\equationautorefname~#1\null{(#1)\null}
\def\figureautorefname~#1\null{Fig. #1\null}

\title{\LARGE \bf

Robust Impedance Control for Dexterous Interaction Using Fractal Impedance Controller with IK-Optimisation
}

\author{Carlo Tiseo, Quentin Rouxel, Zhibin Li, and Michael Mistry 
\thanks{This work has been supported by EPSRC UK RAI Hub ORCA (EP/R026173/1), the Future AI and Robotics for Space (EP/R026092/1), National Centre for Nuclear Robotics (NCNR EPR02572X/1) and THING project in the EU Horizon 2020 (ICT-2017-1).}
\thanks{Carlo Tiseo is with the School of Engineering and Informatics, University of Sussex, UK. All the authors are with the Edinburgh Centre for Robotics, Institute of Perception Action and Behaviour, School of Informatics, University of Edinburgh, UK 
        {\tt\small carlo.tiseo@ed.ac.uk}}%
}

\begin{document}

	\thispagestyle{empty}
\fbox{
\parbox{\textwidth}{
© 2022 IEEE. Personal use of this material is permitted.  Permission from IEEE must be obtained for all other uses, in any current or future media, including reprinting/republishing this material for advertising or promotional purposes, creating new collective works, for resale or redistribution to servers or lists, or reuse of any copyrighted component of this work in other works.}}
\newpage

\maketitle
\thispagestyle{empty}
\pagestyle{empty}

\begin{abstract}
Robust dynamic interactions are required to move robots in daily environments alongside humans. Optimisation and learning methods have been used to mimic and reproduce human movements. However, they are often not robust and their generalisation is limited. This work proposed a hierarchical control architecture for robot manipulators and provided capabilities of reproducing human-like motions during unknown interaction dynamics. Our results show that the reproduced end-effector trajectories can preserve the main characteristics of the initial human motion recorded via a motion capture system, and are robust against external perturbations. The data indicate that some detailed movements are hard to reproduce due to the physical limits of the hardware that cannot reach the same velocity recorded in human movements. Nevertheless, these technical problems can be addressed by using better hardware and our proposed algorithms can still be applied to produce imitated motions. 
\end{abstract}

\section{INTRODUCTION}

Performing dynamic tasks under uncertain interaction is currently one of the main challenges in robotics. Humans and animals are a great source of inspiration for scientists to obtain such performances, considering their ability to handle these conditions \cite{flash2005motor,hogan2012dynamic,hogan2013dynamic,Tiseo2021HapFic,Tiseo2021H-FIC,ijspeert2013dynamical}. A series of optimisation based methods have been proposed to target the problem from a control and a planning perspective. Recently, we have seen the emergence of vertically integrated pipelines to teleoperate robots to reproduce human movements\cite{elobaid2019telexistence,hwangbo2019learning,ravichandar2020recent,penco2018robust,darvish2019whole}. These architectures are often complex, require multi-modal sensing on the user, hardware-specific, and often limited to quasi-static movements. Ideally, a robot should be able to reproduce a movement simply by measuring the position of the demonstrated trajectories in multiple dynamic conditions. Doing so will mimic the ability of humans to learn by imitation while being robust to delays, discretisation, and allowing space-time scaling to adapt the movements to the hardware limitations.

Robust dexterous interaction with variable environments is central for having robot mimicking human movements and performing activities of daily living \cite{elobaid2019telexistence}. In recent years, we have seen multiple examples both in teleoperation and manipulation \cite{wen2020force}. However, these actions are often quasi-static and rely on complex architecture with multiple optimisations and state estimation stages for both the robot and the human \cite{darvish2019whole,penco2018robust,elobaid2019telexistence,pmlr-v78-rana17a}. Another option is to learn a motion from a human to be reused later when needed. Dynamic primitives are an attempt to encode motions through their attractor dynamics to implement learning by demonstration \cite{ijspeert2013dynamical,pmlr-v78-rana17a,atkeson1997robot}. However, this requires performing a regression of the interaction with the environment, which intrinsically couples the system and environmental dynamics \cite{hogan2012dynamic,hogan2013dynamic,Averta2020}. Thus, the learned information is often difficult to generalise and be applied to affine cases (e.g., same action but different interaction dynamics). In contrast, it is known that humans and animals are capable of learning by demonstration even without physical interaction, and they are capable of adapting their behaviour to various dynamic conditions. 

Though recently, data-driven approaches using neural networks have achieved outstanding results, evaluating the quality and completeness of the data-set remains one of the main challenges to certify the safety and robustness of these methods \cite{hwangbo2019learning}. The learning performance is dependent on the quality of the training set, which are large sets of data that are difficult to evaluate, especially in ample and highly non-linear state space of these problems. Consequently, it is challenging to assess the possibility of the system to become unstable or unsafe due to the encounter with a novel scenario not included in the data set. Therefore, the combination of learning algorithms with a robust control architecture might combine the guarantee of the controller with the flexibility of machine learning. 

\begin{figure*}[thb]
      \centering
      \includegraphics[width=.9\textwidth, trim= .5cm 23.5cm .25cm 0cm,clip]{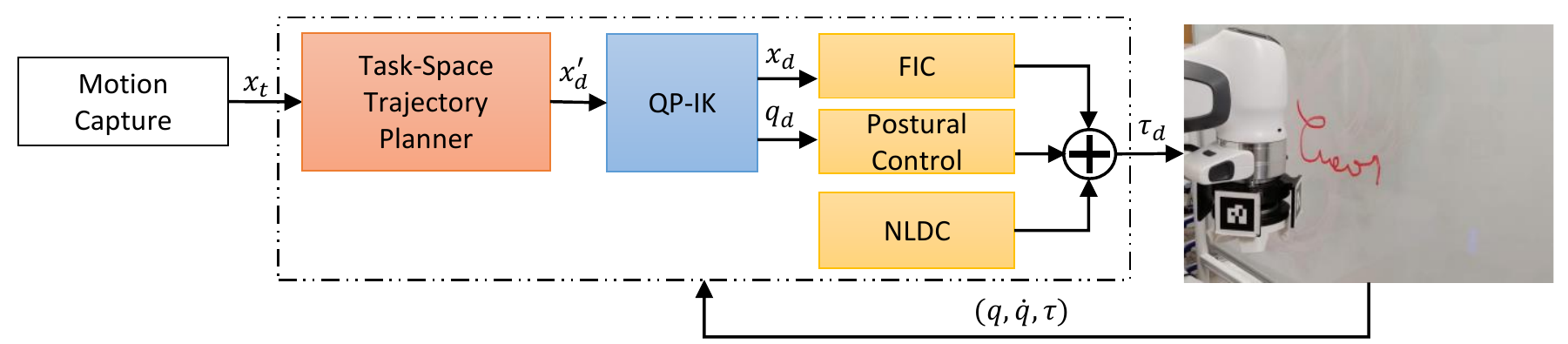}
      \caption{The proposed architecture takes as input a sequence of targets $x_\text{t}$ recorded from a human. The Task-Space Trajectory Planner transforms the stream of targets in the desired trajectory ($x_\text{d}^\prime$). The trajectory is passed to the QP-IK which identifies the optimal posture and trade-off the tracking accuracy to enforce velocity and position limits in joint-space. The output is the desired pose ($x_\text{d}$) and posture ($q_d$), which are passed to the controllers to generate the torque commands. They are then added with the NLDC to determined the desired torques vectors ($\tau_d$). The proposed method also receive in input the measured posture ($q$) and joint velocity ($\dot q$). The torques ($\tau$) are used to estimate the interaction forces, and they are not used in control.}
      \label{fig:Overview}
\end{figure*}
Passive controllers have been studied for more than two decades due to their theoretical robustness to uncertain dynamics and information delays. Recently, the Fractal Impedance Controller (FIC) has been proposed in \cite{babarahmati2019,tiseo2020Planner,tiseo2020}. This controller exploits the introduction of an anisotropic attractor to stabilise a non-linear stiffness in redundant systems without requiring inverse projection or non-linear optimisation. Consequently, the FIC is more robust compared to energy-tank and variable impedance controllers because it does not relies on the path integral (e.g., energy tank) in the control loop \cite{babarahmati2019,raiola2018development} . The method has been successfully applied to generate robust interaction even in the presence of reduced bandwidth and delay in the feedback. Other successful applications have been in trajectory planning in quasi-convex domains, teleoperation, and computational neuroscience \cite{tiseo2021,Tiseo2021H-FIC,babarahmati2020}. Particularly, the FIC allowed the implementation of the Harmonic Passive Motion Paradigm (H-PMP) to model human motor control, which is capable of generalising point to point reaching movements in various dynamic conditions. 

This manuscript exploits some of the components of the H-PMP to formulate a robust controller for robots to track human movements from the recording of the motion trajectories. The method is robust space-time scaling and exploits the benefit of quadratic programming for solving the Inverse Kinematics (QP-IK) problem to trade off the trajectory tracking to respect the mechanical constraint of the hardware in joint space while performing postural optimisation. To summarise the technical contributions made are: 
\begin{enumerate}
    \item We extended the planner part in \cite{tiseo2020Planner}, and applied the vibration theory to autonomously calibrate the friction in the planning, which enables the motion to scale in time. 
    \item We introduced Quadratic Programming (QP) optimisation into the architecture to solve the inverse kinematic with redundancy, which bound the trajectory to the kinematics constraints of the robot with optimised postures. This enables direct streaming of the motion capture trajectories without any post-processing.
\end{enumerate}

The \autoref{sec:method} presents the formulation of the components of the proposed architecture, describes how the human movement are recorded, and describes the experiments. The \autoref{sec:results} describes the experimental set-up and the results of the experiments. \autoref{sec:conclusions} draws the conclusions.

\section{Method} \label{sec:method}
The architecture is composed of an updated version of the H-PMP task-space planner, the QP-Inverse Kinematics (QP-IK) algorithm, Postural Control, Non-Linear Dynamics Compensation (NLDC), and a Task-Space FIC, as shown in \autoref{fig:Overview}. 

\subsection{Task-Space Planner}
The task-space planner take as input the desired trajectory, encoded as next via point ($x_\text{t}$), and desired tangential velocity ($v_\text{d}$). It generalises interactions by extending the harmonic trajectory planner to trajectory tracking by including a viscous field (i.e., $-\mu\dot{x}_\text{d}^\prime$).
This additional viscous field allows us to generalise the point to point planner for continuous trajectory tracking. The mono-dimensional equations are:
\begin{equation}
    \label{eq:HarmonicTrajectoryPlanner}
         x_\text{d}^\prime=\displaystyle{\int_0^t \dot{x}_\text{d}^\prime~dt} 
\end{equation}
\begin{equation}
    \label{eq:HarmonicTrajectoryPlannerV}
        \dot x_{d}^\prime=\displaystyle{\int_0^t \ddot{x}_\text{d}^\prime~dt~~~ \in [-v_\text{d},v_\text{d}]}
\end{equation}
\begin{equation}
    \label{eq:HarmonicTrajectoryPlannerA}
   \ddot{x}_\text{d}^\prime=\left\{\begin{array}{l}
                 \mathrm{sign}\left(\tilde{x}_\text{t}\right)\min\left(\cfrac{K}{M_\text{d}}\left|\tilde{x}_\text{t}\right|,a_\text{max}\right)-\mu\dot{x}_\text{d}^\prime, ~\text{D}\\
                 \cfrac{2A_\text{max}}{\tilde{x}_{\text{T}0}}\left( x_\text{d}^\prime\left(t-1\right)-\cfrac{\tilde{x}_{\text{T}0}}{2}\right)-\mu\dot{x}_\text{d}^\prime,  ~~~~~\text{C}
              \end{array}\right.
\end{equation}
\noindent where $\tilde{x}_\text{t}(t)=x_\text{t}(t)-x_\text{d}^\prime(t-1)$, $a_\text{max}$  describes the maximum acceleration limit of the movement, and $\mu$ is the viscosity. In regards of the equation of the convergence phase (C), $A_\text{max}$ is the acceleration associated with the maximum displacement ($\tilde{x}_{\text{T}0}$) reached in the previous divergence phase (D). The convergence phase is discriminated from the divergence phase based on the evolution of over time the state error $\tilde{x}$, which is decreasing in magnitude without zero crossing (i.e, $|\tilde{x}_\text{t}(t)|<|\tilde{x}_\text{t}(t-1)|$ and $\mathit{sign}(\tilde{x}_\text{t}(t))= \mathit{sign}(\tilde{x}_\text{t}(t-1))$) during convergence. 

Equation \autoref{eq:HarmonicTrajectoryPlannerA} describe an impedance controller for a unit of inertia; thus, the equation describes the equivalent dynamics of a mass-spring-damper system. As a consequence, the controller parameters can be derived from vibration theory setting the desired band-pass for the movements. The following equation is kept in the general vector form to show how it generalises to multiple dimensions because it can not be decoupled as \autoref{eq:HarmonicTrajectoryPlanner}.
\begin{equation}
    \label{eq:HarmonicTrajectoryPlannerParameter}
    \begin{array}{c}
            K=\omega_n^2, ~
            \mu=2 \zeta \omega_n\\
            v_\text{max}=1.595\min\left(v_\text{d},~\omega_n ||\bm{d}|| \right)\\
            a_\text{max}=2\left(\cfrac{v_\text{max}}{||\bm{d}||}\right)^2\bm{d}
    \end{array}
\end{equation}
where $\omega_n=2\pi f_\text{n}$ is the desired natural frequency,  $\zeta$ is the damping ratio, $v_d$ is the desired tangential velocity, and $\bm{d}$ is the distance vector from $x_\text{t}$ when a new target is issued.

It is worth noting that although the proposed method seems not capable of concurrently describing periodic and aperiodic movements. It can be tuned online to switch from oscillatory to non-oscillatory movements by adjusting the value of $\zeta$.
\begin{figure}[thb]
      \centering
      \includegraphics[width=\columnwidth, trim= 5cm 9cm 5cm 9cm,clip]{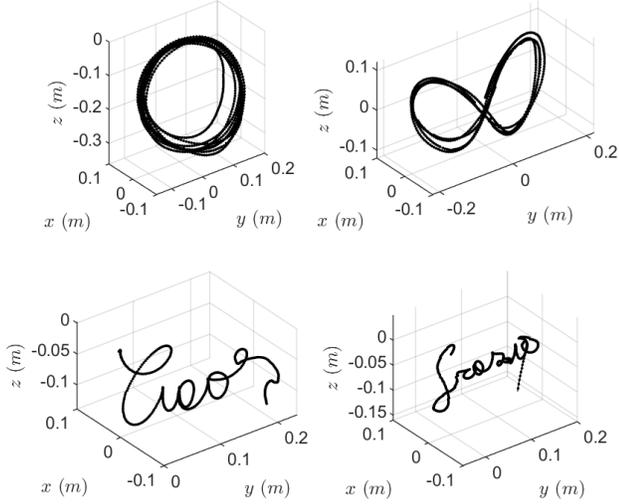}
      \caption{The Circle, the Figure-Eight, `Ciao' and `Grazie' movements recorded from the subject using the motion capture.}
      \label{fig:recordedMovements}
\end{figure}
\subsection{QP-Inverse Kinematics}
The QP-IK algorithms is used to obtain the postural optimisation of the robot while imposing the kinematics constrained associated with the mechanical hardware. This type of solution can handle the exception associate with singularities compared to the projected dynamics used in \cite{Tiseo2021H-FIC} while avoiding the increased architectural complexity associated with the kineto-static duality solution proposed for computational neuroscience in \cite{Tiseo2021H-FIC}.

The QP-IK continuously optimises the postural changes $\Delta \bm{q} \in \R^n$ which is integrated to obtain the desired posture $\bm{q}^d_{t+1} = \bm{q}^d_{t} +\Delta \bm{q} \Delta t$. At each time step $t$, the QP-IK solves the following constrained least-square optimisation:
\begin{equation}
\begin{aligned}\label{QP-IK}
    \min_{\Delta \bm{q}} 
    \norm{\bm{J}(\bm{q}^d_t)\Delta \bm{q}+(\bm{X}_d^\prime \ominus \bm{X}(\bm{q}^d_t))}^2_{\bm{w}_{\text{task}}} +
    \norm{\Delta \bm{q}}^2_{\bm{w}_{\text{reg}}}\\
    \text{s.t.~} \bm{q}_{\text{min}} \leq \Delta \bm{q} + \bm{q}^d_{t} \leq \bm{q}_{\text{max}},
\end{aligned}
\end{equation}
where
$\bm{q}^d_{t} \in \R^n$ is the previous desired joint positions,
$\bm{J}(\bm{q}^d_t) \in \R^{6 \times n}$ is the Jacobian of the end-effector,
$\bm{X}(\bm{q}^d_t) \in \SE$ is the current end-effector pose computed from forward kinematics,
$\bm{X}_d^\prime  \in \SE$ is the desired end-effector pose, 
$\bm{q}_{\text{min}}, \bm{q}_{\text{max}} \in \R^n$ are the minimum and maximum kinematic position limits for each joint,
$\bm{w}_{\text{task}}, \bm{w}_{\text{reg}} \in \R$ are the task and regularisation weights tuned for stability,
and $\ominus$ is the subtraction operator defined for the Lie algebra of $\SE$.

\subsection{Monodimensional Fractal Impedance Controller}
The task-space FIC is defined based on the following force profile for every dimension.
\begin{equation}
    \label{eq:FICForce}
    \begin{array}{l}
         F_\text{c}=\left\{\begin{array}{ll}K_0\tilde{x}=
              K_0 \left(x_\text{d}-x\right),& |\tilde{x}|\le \xi \tilde{x}_\text{b} \\
              \cfrac{\Delta F}{2}\left(\tanh\left(\cfrac{\tilde{x}-\tilde{x}_\text{b}}{S\tilde{x}_\text{b}}+\pi\right)+1\right)+&\\+F_0,& \text{else}
              \end{array}\right.\\
    \end{array}
\end{equation}
where $K_0$ is the constant stiffness, $\tilde{x}=x_\text{d}-x$ is the end-effector pose error, $\Delta{F}=F_\text{max}-F_0$, $F_0=\xi K_0\tilde{x}_\text{b}$, $S=\left(1- \xi
\right)\left(\tilde{x}_\text{b}/(2\pi)\right)$ controls the saturation speed, and $\xi \in [0,~1]$ control the starting of the saturation behaviour while approaching the state error magnitude of $\tilde{x}_\text{b}$ in each direction. The $F_\text{c}$ is then substituted in the attractor formulation.
\begin{equation}
    \label{eqFICattractor}
    F(\tilde{x})=\left\{\begin{array}{cc}
        F_\text{c}(\tilde{x}),  & \text{D} \\
        \cfrac{2F_\text{c}(\tilde{x}_\text{max})}{\tilde{x}_\text{max}}\left(\tilde{x}-\cfrac{\tilde{x}_\text{max}}{2}\right) & \text{C}
    \end{array}\right.
\end{equation}
where $\tilde{x}_\text{max}$ is the maximum state error recorded at the beginning of the last convergence phase.

\subsection{Torque Command}
The control command in torque is generated by superimposing the FIC task-space controller, the postural controller, and the compensation for the robot's non-linear dynamics ($C(q,\dot{q})$) and gravitational forces ($G(q)$).

\begin{equation}
    \label{eq:TorqueCmd}
    \tau_\text{d}=C(q,\dot{q})+G(q)+ K_\text{JS} \tilde{q} - D_\text{JS} \dot{q} +J_\text{ee}^T(W_\text{FIC}-D_\text{TS}\dot{X})
\end{equation}
where $q$ is the joint state, $\tilde{q}$ is the joint state error, $K_\text{JS}$ is the postural control stiffness, $D_\text{JS}$ is the postural control viscosity, $J_{ee} \in \mathbb{R}^{6\times n}$ is the end-effector jacobian, $W_\text{FIC} \in \mathbb{R}^{6}$ is the FIC wrench, $D_\text{TS} \in \mathbb{R}^{6\times6}$ is the Task-Space viscosity, and $\dot{X} \in \mathbb{R}^{6}$ is the end-effector velocity.

\subsection{Space-time Trajectory Scaling}
The space scaling is performed by multiplying the point coordinates by a scaling factor $S_\text{x} \in \R$. The time scaling is obtained by multiplying the sampling frequency by the scaling factor $S_\text{t} \in \R$. Given a recorded human trajectory $x_\text{h}(t)$ recorded in the interval $[0,t_\text{f}]$, using the sampling frequency $f_\text{s}$. The scaled trajectory is:
\begin{equation}
    \label{eq:scaling}
    x_\text{t}(t^\prime)=S_\text{x}x_\text{h}(kS_\text{t}^{-1}f_\text{s}^{-1}),~t^\prime \in [0,~t_\text{f}S_\text{t}^{-1}]
\end{equation}
where $k\in[0,~f_\text{s}t_\text{f}] \subset \mathbb{N}$ selects the available data sample.
The tangential velocity is computed online with finite-difference over the scaled demonstrated trajectory. The four trajectories recorded from the subject are shown in \autoref{fig:recordedMovements}.

\section{Experimental Validation}\label{sec:results}

\subsection{Motion Recording}
The writing movements from the subject were recorded using a Vicon Motion Capture System. The hand-movement were tracked using the rigid-body tracking tool, placing five markers on the subject's hand at a frequency of \SI{100}{\hertz}. Two markers were placed on the $1^{st}$ and $4^{th}$ metatarsal joints; one marker was placed in the centre of the metatarsus, a marker was on the Scaphoid bone, and the last was on the Pisiform bone. The recorded trajectories were exported in comma-separated values (CSV) file format, which is loaded and streamed online to the proposed controller.

\subsection{Experimental Set-Up}
The validation of the proposed method has been conducted in three stages. 
At the beginning we evaluated the proposed planner against a minimum velocity bang-bang planner in both velocity and tracking accuracy using multiple sets of parameters (\autoref{tab:parameters}) to write `Ciao'. The evaluation has been done comparing the RMSE from the input trajectory for the position and velocity on the motion plane. 

The second test is conducted on the robot to test the accuracy of the system without interaction, and how the tracking is affected by the tuning of the parameters. The writing tasks are the capital \textit{B}, \textit{F} and \textit{H} at a speed equal to 1/4 of the recorded human trajectories. The following two sets of parameters are used: Set 1 is $\tilde{x}_\text{b}=\SI{.05}{\meter}$, $K_0=\SI{200}{\newton\meter^{-1}}$ and $F_\text{max}=\SI{20}{\newton}$, and Set 2 is  $\tilde{x}_\text{b}=\SI{.02}{\meter}$, $K_0=\SI{1200}{\newton\meter^{-1}}$ and $F_\text{max}=\SI{30}{\newton}$.

The last test is drawing the trajectories in \autoref{fig:recordedMovements} on a whiteboard. The Circle and Figure-Eight trajectories are also tested with external perturbations, which have been introduced either by pushing the robot directly or moving the board. This latter experiment evaluates the evaluation of robot compliant behaviour, as well as its robustness.
\begin{table}[h]
    \centering
    \begin{tabular}{c| c c c c}
         N & $\zeta$   & $f_\text{n}$ & $a_\text{max}$ & $v_\text{max}$ \\ \hline
         1 &$.005$&   $4$          &     $10$          &    $.3$     \\
         2 &$.010$&   $10$          &    $10$          &    $.4$     \\
         3 &$.010$&   $5$          &     $5$           &    $.3$     \\
         4 &$.050$&   $4$          &     $5$           &    $.3$      \\
         5 &$.050$&   $10$         &     $2$           &    $.4$      \\
         6 &$.100$&   $4$          &     $3$           &    $.3$      \\
    \end{tabular}
    \caption{Sets of parameters used for the planner comparisons. $\zeta$ and $f_\text{n}$ are used only for the proposed method.  $a_\text{max}$ ($\si{\meter\per\second^2}$) is used for planners. $v_\text{max}$ ($\si{\meter\per\second^2}$) is used only from the bang-bang trajectory planner.}
    \label{tab:parameters}
\end{table}
\subsection{Results}
\begin{figure}[thb]
      \centering
      \includegraphics[width=\columnwidth, trim= 3.75cm 10.25cm 3.15cm 9.75cm,clip]{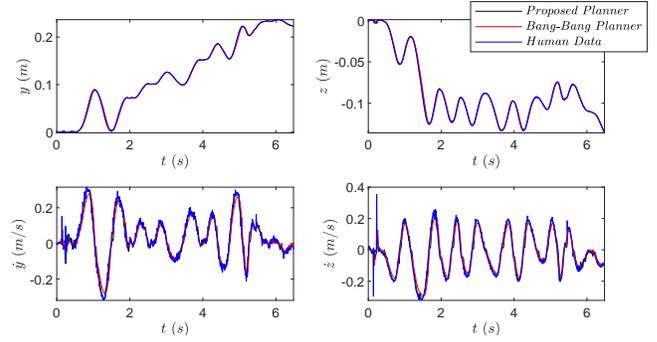}
      \caption{Comparing the human data and the trajectories generated by the two planners using the third set of parameters ( N$=3$ ) shows a negligible difference between the three trajectories. However, the velocity profiles show that the velocity profile of the bang-bang planner less accurate than the proposed method. Thus, confirming the comparison between the values reported in \autoref{tab:resultsPP} and \autoref{tab:resultsBBP}}
      \label{fig:CompPosVelParSet3}
\end{figure}

The RMSE for the comparison between the two planners are reported in \autoref{tab:resultsPP} for the proposed approach, and in \autoref{tab:resultsBBP} for the bang-bang planner.  The data show that the proposed method has a better tracking accuracy, and it is more robust to selecting sub-optimal parameters. For example, let's consider the third set of parameters, the best tracking performance for both planners. It shows that our planner is two times more accurate in position, and it is about \SI{20}{\percent} better in tracking the reference velocity. However, \autoref{fig:CompPosVelParSet3} shows that the trajectories do not have a sensible distance in position. However, there is a small deviation of the bang-bang velocity profile from both the proposed planner and the human data. It is also worth noting that both planners are computationally inexpensive, scale easily, and can both be deployed online.

\begin{figure}[thb]
      \centering
      \includegraphics[width=\columnwidth, trim= 6cm 11.25cm 6cm 10.65cm,clip]{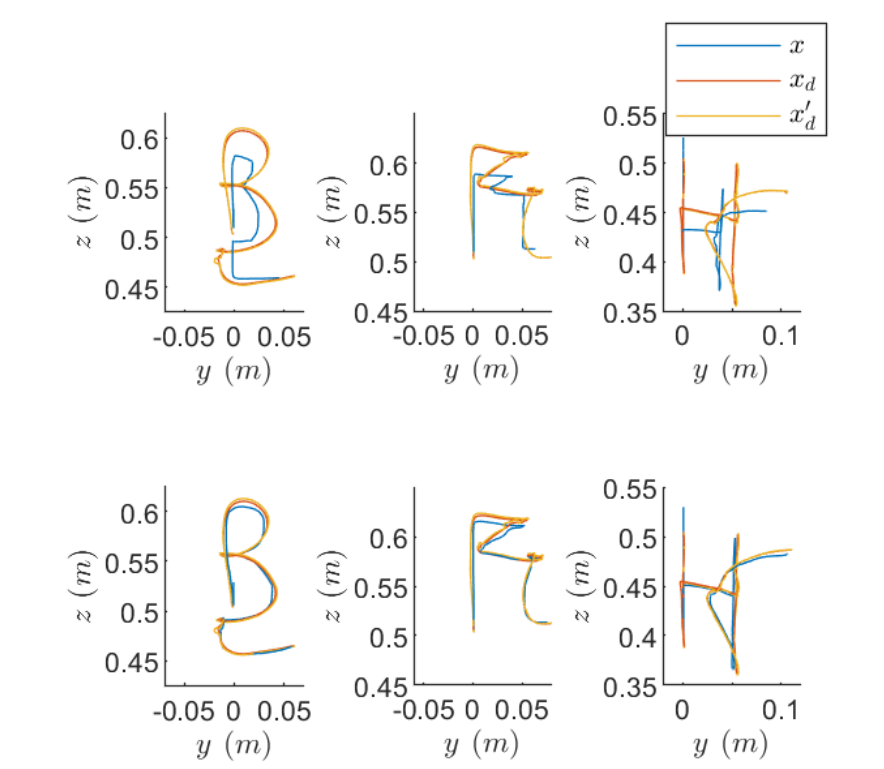}
      \caption{The letter written with the first set of parameters ($x_\text{b}=\SI{.05}{\meter}$, $K_0=\SI{200}{\newton\meter^{-1}}$ and $F_\text{max}=\SI{20}{\newton}$) are on the top row. The letters written with the second set ($x_\text{b}=\SI{.02}{\meter}$, $K_0=\SI{1200}{\newton\meter^{-1}}$ and $F_\text{max}=\SI{30}{\newton}$) are on the bottom. As can be expected an higher rigidity of the controller allows better tracking of the desired trajectory.}
      \label{fig:SingleLetterComparison}
\end{figure}

\begin{figure}[thb]
      \centering
      \includegraphics[width=\columnwidth, trim= 4cm 12cm 4cm 12cm,clip]{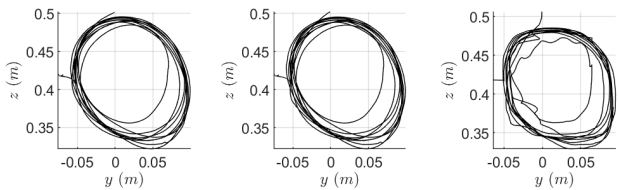}
      \caption{The Circle trajectory scaled in speed and dimension is shown. The trajectory generated from the planner is on the leftmost. The trajectory mediated by the QP-IK is in the centre. The trajectory executed from the robot is on the rightmost.}
      \label{fig:CircleScaling}
\end{figure}

\begin{table}[ht]
    \centering
    \begin{tabular}{c| c c c c}
         N & $\mathrm{RMSE}(y)$   & $\mathrm{RMSE}(z)$ & $\mathrm{RMSE}(\dot{y})$   & $\mathrm{RMSE}(\dot{z})$ \\ \hline
         1 &$.002$&$.002$&$.012$&$.016$\\
         2 &$.001$&$.001$&$.022$&$.023$\\
         3 &$.001$&$.001$&$.013$&$.016$\\
         4 &$.002$&$.002$&$.012$&$.017$\\
         5 &$.001$&$\approx0$&$.021$&$.022$\\
         6 &$.002$&$0.002$&$.012$&$.017$\\
    \end{tabular}
    \caption{Results for the planner comparisons. The positions are in $\si{\meter}$ and the velocities in $\si{\meter\per\second^2}$.}
    \label{tab:resultsPP}
\end{table}
The trajectories obtained for the letter \textit{B}, \textit{F} and \textit{H} with the two sets of parameters are shown \autoref{fig:SingleLetterComparison}. The data show that despite the more compliant controllers retains the geometrical features; it shrinks the letter dimension compared to the desired dimension. It is worth reminding that due to special characteristics of the FIC controller, the parameters can be updated online without any risk for the system stability \cite{babarahmati2019,tiseo2020Planner,Tiseo2021HapFic}. The space-time scalability of the trajectories has been tested on the Circle, which has been shrunk (\autoref{fig:CircleScaling}) and slowed down, as can be appreciated in the attached video. It is also worth noting that all the motion played on the robot reached up to half of the human movement speed due to hardware limitations. 
\begin{table}[ht]
    \centering
    \begin{tabular}{c| c c c c}
         N & $\mathrm{RMSE}(y)$   & $\mathrm{RMSE}(z)$ & $\mathrm{RMSE}(\dot{y})$   & $\mathrm{RMSE}(\dot{z})$ \\ \hline
         1 &$.001$&$.001$&$.025$&$.026$\\
         2 &$.001$&$.001$&$.025$&$.027$\\
         3 &$.002$&$.002$&$.018$&$.017$\\
         4 &$.002$&$.002$&$.018$&$.017$\\
         5 &$.007$&$.007$&$.048$&$.049$\\
         6 &$.004$&$.004$&$.024$&$.022$\\
    \end{tabular}
    \caption{Results for the bang-bang planner. The positions are in $\si{\meter}$ and the velocities in $\si{\meter\per\second^2}$.}
    \label{tab:resultsBBP}
\end{table}
\begin{figure}[thb]
      \centering
      \includegraphics[width=\columnwidth, trim= 3.5cm 10cm 4cm 10cm,clip]{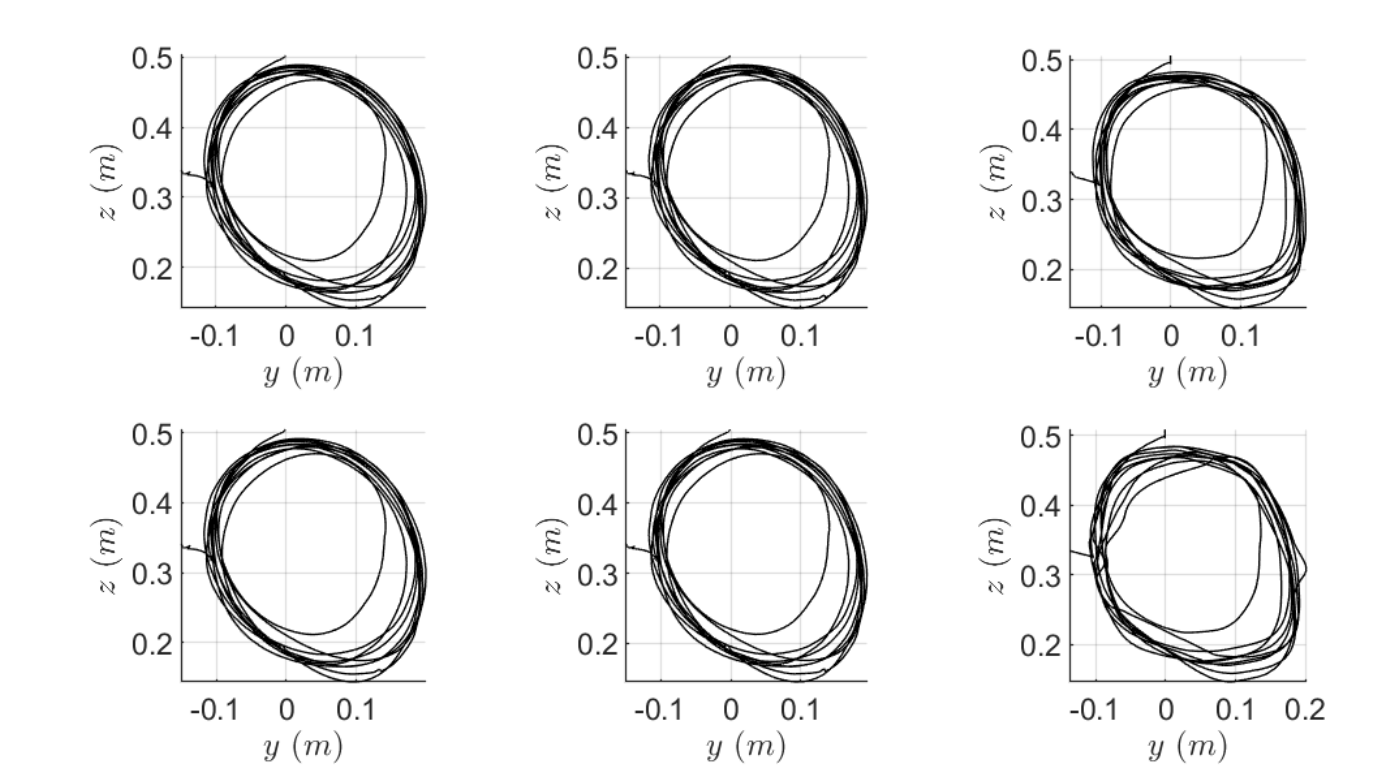}
      \caption{The Circle performed without perturbation is on the top and the Circle performed with perturbation is on the bottom. The trajectories generated from the planners are on the leftmost. The trajectories mediated by the QP-IK are in the centre. The trajectories executed from the robot are on the rightmost.}
      \label{fig:CirclePertUnpert}
\end{figure}
The comparison between the unperturbed and perturbed conditions for the Circle and the Figure-Eight are shown in \autoref{fig:CirclePertUnpert} and \autoref{fig:Figure8PertUnpert}, respectively. The magnitudes of the interaction forces recorded in the experiments are reported in \autoref{fig:ForcesCircleFig8} for the Circle and the Figure-Eight. The trajectories at the different stages of the proposed method for `Ciao' and `Grazie' are shown in \autoref{fig:CiaoGrazie}, while the magnitude of interaction forces are shown in \autoref{fig:ForcesCiaoGrazie}. The data indicate that the tracking error is introduced at the hardware level. Nevertheless, the words still retain their main features despite the robot does not allow to fully capture all the feature of these words \autoref{fig:CiaoGrazie}, validating how the proposed method can preserve the main characteristics of complex movements. However, the more intricate details are lost at the controller level during writing. This indicates that the chosen controller parameters might require some \textit{ad hoc} tuning in this tasks. Therefore, considering that all the parameters of the proposed architecture can be updated online, a data-driven approach using machine learning can be considered a possible solution to this limitation. 
\begin{figure}[thb]
      \centering
      \includegraphics[width=\columnwidth, trim= 3.5cm 10cm 4cm 10cm,clip]{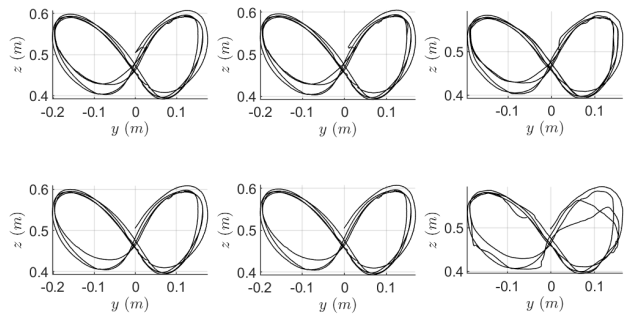}
      \caption{The Figure-Eight performed without perturbation is on the top and with perturbation is on the bottom. The trajectories generated from the planners are on the leftmost. The trajectories mediated by the QP-IK are in the centre. The trajectories executed from the robot are on the rightmost.}
      \label{fig:Figure8PertUnpert}
\end{figure}
\begin{figure}[thb]
      \centering
      \includegraphics[width=\columnwidth, trim= 6cm 10cm 6cm 10cm,clip]{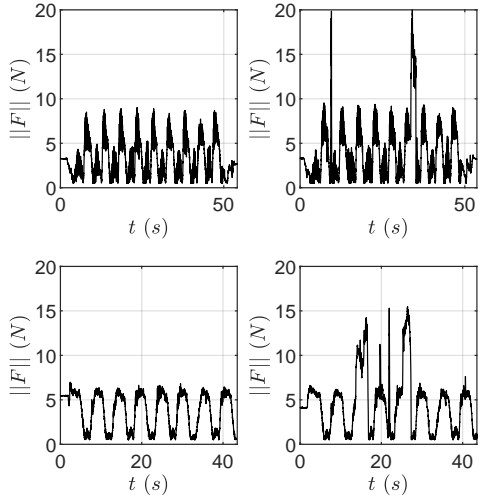}
      \caption{The magnitudes of the interaction forces during unperturbed experiments are on the left, perturbed on the right. The Circle's data are on the first row, and the data for the Figure-Eight are on the second row.}
      \label{fig:ForcesCircleFig8}
\end{figure}
\begin{figure}[thb]
      \centering
      \includegraphics[width=\columnwidth, trim= 3.5cm 10cm 4cm 10cm,clip]{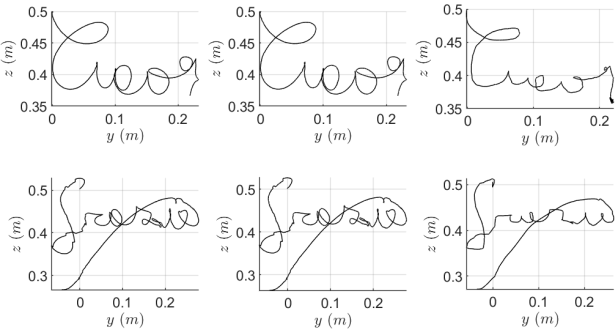}
      \caption{`Ciao' and `Grazie' trajectories performed without external perturbations. The trajectories generated from the planners are on the leftmost. The trajectories mediated by the QP-IK are in the centre. The trajectories executed from the robot are on the rightmost.}
      \label{fig:CiaoGrazie}
\end{figure}
\begin{figure}[thb]
      \centering
      \includegraphics[width=\columnwidth, trim= 6cm 12cm 6cm 12cm,clip]{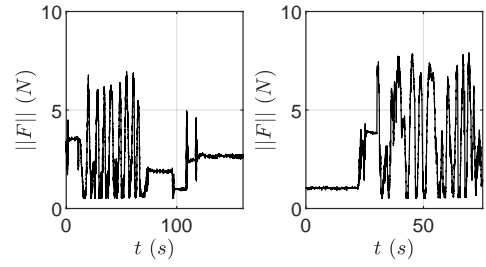}
      \caption{The figure reports the magnitudes of the interaction forces recorded during the two writing experiments. The `Ciao data are on the left, and `Grazie' data are on the right.}
      \label{fig:ForcesCiaoGrazie}
\end{figure}
\section{Conclusions}\label{sec:conclusions}
The introduction of the QP-IK simplifies the deployment of the H-PMP to robots by reducing the numbers of parameters to be tuned. However, this comes at the cost of reduced robustness to singularity, similarly to a null-space controller relying on projected dynamics. Nevertheless, the QP-IK discourages the convergence to singular configuration, making it more robust than the projected dynamics. Regarding applications where singular configuration might be helpful in some task (e.g., legged locomotion). Singular configurations can be handled as exceptions (i.e., specific target postures), or the H-PMP solution to the postural control can be implemented. The results show that the proposed method can replicate a human-like trajectory without the need for any learning or model-based control of the robot interaction. This allows the robot to be robust to external perturbation and be able to make and break contact without any stability concerns. The robot accuracy in tracking and recovering from the introduction of external perturbation is determined by the FIC non-linear impedance, which adapts the stiffness based on the pose error. This characteristic is also what allows the system to be accurate to model errors which can be regarded as external perturbations. It is worth mentioning that the compensation of the non-linear dynamics can be removed \cite{tiseo2020}. It has been introduced to reduce the load on the FIC spring, which makes the system more responsive. Furthermore, the ability to tune the parameters online and the robustness to unknown dynamics sets aside this method from dynamic primitives and neural network approaches. Machine learning methods encode motor strategies that couple together the system and the environmental interaction during the regression process performed the encoding of the motor strategies (i.e., learning). Furthermore, it is a challenge to guarantee stability and generality in data-driven approaches due to the difficulty of assessing the quality of large and complex data sets. On the other hand, the proposed approach has a strong guarantee of stability and generality, which decouples the tracking and postural optimisation from the system stability. Thus, combining the proposed method with machine learning algorithms might result in robust systems capable of better knowledge generalisation. However, this is just a hypothesis at this stage and requires further investigation.

In summary, the presented an architecture is capable of reproducing human-like movements during robust interaction with the environment. There are still some improvements to be made in reproducing the finer details of complex movements. Nevertheless, these imperfection are mainly related to the need for adaptive tuning of the controller and planner parameters, which could be implemented with machine learning algorithms.







\bibliographystyle{IEEEtran}
\bibliography{root}

\end{document}